\documentclass[twoside]{article}

\usepackage[utf8]{inputenc} 
\usepackage[T1]{fontenc}
\usepackage{amssymb}
\usepackage{amsfonts}

\usepackage{tgtermes}
\usepackage{amsmath}
\usepackage{scalefnt,letltxmacro}
\LetLtxMacro{\oldtextsc}{\textsc}
\renewcommand{\textsc}[1]{\oldtextsc{\scalefont{1.10}#1}}
\usepackage[scaled=0.92]{PTSans}

\usepackage[usenames,dvipsnames]{xcolor}
\definecolor{shadecolor}{gray}{0.9}

\usepackage[parfill]{parskip}
\usepackage{afterpage}
\usepackage{framed}
\usepackage{nicefrac}

\usepackage[colorinlistoftodos,
           prependcaption,
           textsize=small,
           backgroundcolor=yellow,
           linecolor=lightgray,
           bordercolor=lightgray]{todonotes}

\usepackage{lineno}

\usepackage{ragged2e}

\DeclareRobustCommand{\parhead}[1]{\textbf{#1}~}

\usepackage{graphicx}
\usepackage[labelfont=it,labelsep=period]{caption}
\usepackage[format=hang]{subcaption}

\usepackage{booktabs}
\usepackage{arydshln} 

\usepackage{natbib}

\usepackage{algorithm}
\usepackage{algorithmic}
\usepackage{listings}
\usepackage{fancyvrb}
\fvset{fontsize=\normalsize}

\usepackage[colorlinks,linktoc=all]{hyperref}
\usepackage[all]{hypcap}
\hypersetup{citecolor=Violet}
\hypersetup{linkcolor=black}
\hypersetup{urlcolor=MidnightBlue}
\usepackage{url}

\usepackage[nameinlink]{cleveref}
\creflabelformat{equation}{#1#2#3}
\crefname{equation}{eq.}{eqs.}  
\Crefname{equation}{Eq.}{Eqs.}

\usepackage[acronym,smallcaps,nowarn]{glossaries}

\usepackage{listings}
\lstdefinestyle{alp_style}{
    commentstyle=\color{OliveGreen},
    numberstyle=\tiny\color{black!60},
    stringstyle=\color{BrickRed},
    basicstyle=\ttfamily\scriptsize,
    breakatwhitespace=false,
    breaklines=true,
    captionpos=b,
    keepspaces=true,
    numbers=none,
    numbersep=5pt,
    showspaces=false,
    showstringspaces=false,
    showtabs=false,
    tabsize=2
}
\DeclareRobustCommand{\E}[2]{\mathbb{E}_{#1}\left[#2\right]}

\DeclareRobustCommand{\diag}[1]{\textrm{diag}\left(#1\right)}

\newcommand{\g}{\, | \,}
\newcommand{\prm}{\, ; \,}

\newcommand{\Lcal}{\mathcal{L}}
\newcommand{\Ncal}{\mathcal{N}}

\newacronym{ADVI}{advi}{automatic differentiation variational inference}
\newacronym{AR}{a{\small\&}r}{augment and reduce}

\newacronym{BBVI}{bbvi}{black-box variational inference}

\newacronym{CDF}{cdf}{cumulative distribution function}
\newacronym{CS-EFE}{cs-efe}{context selection for exponential family embeddings}
\newacronym{CTM}{ctm}{correlated topic model}

\newacronym[\glslongpluralkey={deep exponential families}]{DEF}{def}{deep exponential family}
\newacronym{DMIS}{dmis}{deterministic multiple importance sampling}

\newacronym{EFE}{efe}{exponential family embeddings}
\newacronym{ELBO}{elbo}{evidence lower bound}
\newacronym{EM}{em}{expectation maximization}

\newacronym{GNTS}{gn-ts}{gamma-normal time series model}
\newacronym{G-REP}{g-rep}{generalized reparameterization}

\newacronym{HMC}{hmc}{{H}amiltonian {M}onte {C}arlo}

\newacronym{KL}{kl}{{K}ullback-{L}eibler}

\newacronym{LDA}{lda}{latent {D}irichlet allocation}

\newacronym{MAP}{map}{\emph{maximum a posteriori}}
\newacronym{MCMC}{mcmc}{{M}arkov chain {M}onte {C}arlo}
\newacronym{MF}{mf}{matrix factorization}
\newacronym{MIS}{mis}{multiple importance sampling}

\newacronym{OBBVI}{o-bbvi}{overdispersed black-box variational inference}
\newacronym{OVE}{ove}{one-vs-each}

\newacronym{SIVI}{sivi}{semi-implicit variational inference}
\newacronym{SVI}{svi}{stochastic variational inference}

\newacronym{USIVI}{uivi}{unbiased implicit variational inference}

\newacronym{VAE}{vae}{variational autoencoder}
\newacronym{VEM}{vem}{variational expectation maximization}
\newacronym{VI}{vi}{variational inference}

\usepackage{lipsum}
\usepackage[accepted]{aistats2018mod}
\hypersetup{citecolor=Violet}
\hypersetup{linkcolor=black}
\hypersetup{urlcolor=MidnightBlue}

\begin{document}

\twocolumn[

\aistatstitle{Unbiased Implicit Variational Inference}

\aistatsauthor{ Michalis K.~Titsias \And Francisco J.~R.~Ruiz }

\aistatsaddress{ Athens University of Economics and Business \And  University of Cambridge \& Columbia University } ]

\begin{abstract}
  We develop \emph{\acrlong{USIVI}} (\acrshort{USIVI}), a method that expands the applicability of variational inference by defining an expressive variational family. \acrshort{USIVI} considers an implicit variational distribution obtained in a hierarchical manner using a simple reparameterizable distribution whose variational parameters are defined by arbitrarily flexible deep neural networks. Unlike previous works, \acrshort{USIVI} directly optimizes the \acrfull{ELBO} rather than an approximation to the \acrshort{ELBO}. We demonstrate \acrshort{USIVI} on several models, including Bayesian multinomial logistic regression and variational autoencoders, and show that \acrshort{USIVI} achieves both tighter \acrshort{ELBO} and better predictive performance than existing approaches at a similar computational cost.
\end{abstract}

\section{INTRODUCTION}
\label{sec:introduction}
\glsresetall

\Gls{VI} is an approximate Bayesian inference technique that recasts inference as an optimization problem \citep{Jordan1999learn,Wainwright2008,Blei2017}. The goal of \gls{VI} is to approximate the posterior $p(z\g x)$ of a given probabilistic model $p(x,z)$, where $x$ denotes the data and $z$ stands for the latent variables. \gls{VI} posits a parameterized family of distributions $q_{\theta}(z)$ and then minimizes the \gls{KL} divergence between  the approximating distribution $q_{\theta}(z)$ and the exact posterior $p(z\g x)$. This minimization is equivalent to maximizing the \gls{ELBO}, which is a function $\Lcal(\theta)$ expressed as an expectation over the variational distribution,
\begin{equation}\label{eq:elbo}
	\Lcal(\theta) = \E{q_{\theta}(z)}{\log p(x,z) - \log q_{\theta}(z)}.
\end{equation}
Thus, \gls{VI} maximizes \Cref{eq:elbo}, which involves the log-joint $\log p(x,z)$ rather than the intractable posterior.

Classical \gls{VI} relies on the assumption that the expectations in \Cref{eq:elbo} are tractable and applies a coordinate-wise ascent algorithm to find $\theta$ \citep{Ghahramani2001}. In general, this assumption requires two conditions: the model must be conditionally conjugate (meaning that the conditionals $p(z_n\g x, z_{\neg n})$ are in the same exponential family as the prior $p(z_n)$ for each latent variable $z_n$), and the variational family must have a simplified form such as to be factorized across latent variables (mean-field \gls{VI}).

The above two restrictive conditions when applying \gls{VI} have motivated several lines of research to expand the use of \gls{VI} to more complex settings. To address the conjugacy condition on the model, black-box \gls{VI} methods have been developed, allowing \gls{VI} to be applied on a broad class of models by using Monte Carlo estimators of the gradient \citep{Carbonetto2009,Paisley2012,Ranganath2014,Kingma2014,Rezende2014,Titsias2014_doubly,Kucukelbir2015,Kucukelbir2017}. To address the simplified (typically mean-field) form of the variational family, more complex variational families have been proposed that incorporate some structure among the latent variables \citep{Jaakkola1998,Saul1996,Giordano2015,Tran2015,Tran2016,Ranganath2016,Han2016,Maaloe2016}. See also \citet{Zhang2017} for a review on recent advances on variational inference.

Here, we focus on implicit \gls{VI} where the variational distribution $q_{\theta}(z)$ can have arbitrarily flexible forms constructed using neural network mappings. A distribution $q_{\theta}(z)$ is \emph{implicit} when it is not possible to evaluate its density but it is possible to draw samples from it. One typical way to draw from an implicit distribution in \gls{VI} is to first sample a noise vector and then push it through a deep neural network \citep{Mohamed2016,Huszar2017,Tran2017,Li2018,Mescheder2017,Shi2018}. Implicit \gls{VI} expands the variational family making $q_{\theta}(z)$ more expressive, but computing $\log q_{\theta}(z)$ in \Cref{eq:elbo}---or its gradient---becomes intractable. To address that, implicit \gls{VI} typically relies on density ratio estimation. However, density ratio estimation is challenging in high-dimensional settings. To avoid density ratio estimation, \citet{Yin2018} proposed \gls{SIVI}, an approach that obtains the variational distribution $q_{\theta}(z)$ by mixing the variational parameter with an implicit distribution. Exploiting this definition of $q_{\theta}(z)$, \gls{SIVI} optimizes a sequence of lower (or upper) bounds on the \gls{ELBO} that eventually converge to \Cref{eq:elbo}.

In this paper, we develop an unbiased estimator of the gradient of the \gls{ELBO} that avoids density ratio estimation. Our approach builds on \gls{SIVI} in that we also define the variational distribution by mixing the variational parameter with an implicit distribution. In contrast to \gls{SIVI}, we propose an unbiased optimization method that directly maximizes the \gls{ELBO} rather than a bound. We call our method \textit{\acrlong{USIVI}} (\acrshort{USIVI}). We show experimentally that \acrshort{USIVI} can achieve better \gls{ELBO} and predictive log-likelihood than \gls{SIVI} at a similar computational cost.

We develop \acrshort{USIVI} using a semi-implicit variational approximation $q_{\theta}(z)=\int q_{\theta}(z\g \varepsilon) q(\varepsilon) d\varepsilon$, such that the conditional $q_{\theta}(z\g \varepsilon)$ is a reparameterizable distribution. The dependence of the conditional $q_{\theta}(z\g \varepsilon)$ on the random variable $\varepsilon$ can be arbitrarily complex. We use a deep neural network parameterized by $\theta$ that takes $\varepsilon$ as input and outputs the parameters of the conditional $q_{\theta}(z\g \varepsilon)$. Given $\varepsilon$, the conditional is a ``simple'' reparameterizable distribution; however marginalizing out $\varepsilon$ results in an implicit and more complex distribution $q_{\theta}(z)$. Exploiting these properties of the variational distribution, \acrshort{USIVI} expresses the gradient of the \gls{ELBO} in \Cref{eq:elbo} as an expectation, allowing us to construct an unbiased Monte Carlo estimator. The resulting estimator requires samples from the conditional distribution $q_{\theta}(\varepsilon\g z) \propto q_{\theta}(z\g \varepsilon) q(\varepsilon)$. We develop a computationally efficient way to draw samples from this conditional using a fast \gls{MCMC}   procedure that starts from the stationary distribution. In this way, we avoid the time-consuming burn-in or transient phase that characterizes \gls{MCMC} methods.

\section{UNBIASED IMPLICIT VARIATIONAL INFERENCE}
\label{sec:method}
\glsresetall

In this section we present \gls{USIVI}. First, in \Cref{sec:sivi} we describe how to build the variational distribution, following \gls{SIVI} \citep{Yin2018}. Second, in \Cref{sec:usivi} we show how to form an unbiased estimator of the gradient of the \gls{ELBO}. Finally, in \Cref{sec:algorithm} we put forward the resulting \gls{USIVI} algorithm and explain how to run it efficiently.

\subsection{Semi-Implicit Variational Distribution}
\label{sec:sivi}

To approximate the posterior $p(z\g x)$ of a probabilistic model $p(x,z)$, \gls{USIVI} uses a semi-implicit variational distribution $q_{\theta}(z)$ \citep{Yin2018}. This means that $q_{\theta}(z)$ is defined in a hierarchical manner with a mixing parameter,
\begin{equation}\label{eq:implicit_dist_samp}
	\varepsilon\sim q(\varepsilon), \quad
	z\sim q_{\theta}(z\g \varepsilon),
\end{equation}
or equivalently,
\begin{equation}\label{eq:implicit_dist_int}
	q_{\theta}(z)=\int q_{\theta}(z\g \varepsilon) q(\varepsilon) d\varepsilon. 
\end{equation}
\Cref{eq:implicit_dist_samp,eq:implicit_dist_int} reveal why the resulting variational distribution $q_{\theta}(z)$ is implicit, as we can obtain samples from it (\Cref{eq:implicit_dist_samp}) but cannot evaluate its density, as the integral in \Cref{eq:implicit_dist_int} is intractable.\footnote{This is similar to a hierarchical variational model \citep{Ranganath2016}; however we do not assume that the conditional $q_{\theta}(z\g \varepsilon)$ can be factorized. See \citet{Yin2018} for a discussion on the differences between hierarchical variational models and \gls{SIVI}.}

The dependence of the conditional $q_{\theta}(z\g \varepsilon)$ on the random variable $\varepsilon$ can be arbitrarily complex. In \gls{USIVI}, its parameters are the output of a deep neural network (parameterized by the variational parameters $\theta$) that takes $\varepsilon$ as input.

\begin{figure}[t]
	\centering
	\includegraphics[width=0.33\textwidth]{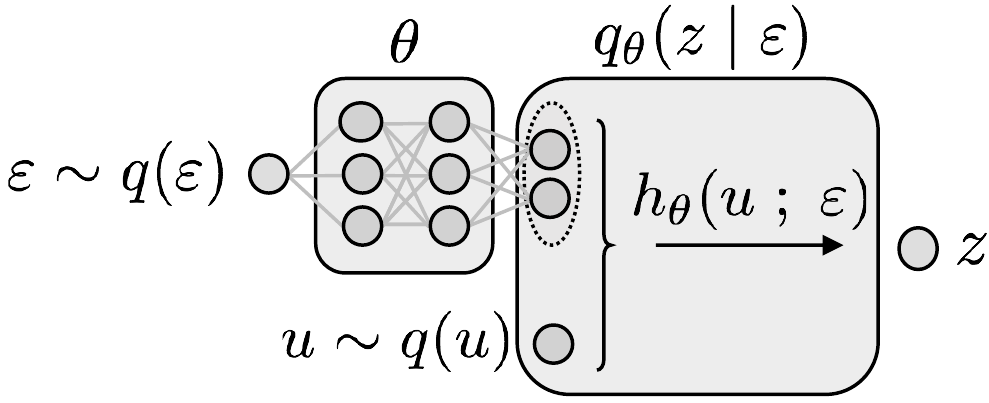}
	\caption{Illustration of the sampling procedure for the implicit variational distribution $q_{\theta}(z)$. First, a sample $\varepsilon\sim q(\varepsilon)$ is pushed through a neural network with parameters $\theta$ (left block). This network outputs the parameters (dotted ellipse) of the conditional distribution $q_{\theta}(z\g\varepsilon)$. Since the conditional is reparameterizable (right block), to draw a sample $z$ we can first sample $u\sim q(u)$ and then set $z=h_{\theta}(u\prm \varepsilon)$, where $h_{\theta}(\cdot)$ is an appropriate transformation. The transformation $h_{\theta}(\cdot)$ depends on $\varepsilon$ and $\theta$ through the parameters of the conditional. The output $z=h_{\theta}(u\prm \varepsilon)$ is a sample from the variational distribution $q_{\theta}(z)$.\label{fig:semi_implicit_illustration}}
\end{figure}

\parhead{Assumptions.}
In \gls{USIVI}, the conditional $q_{\theta}(z\g \varepsilon)$ must satisfy two assumptions. First, it must be reparameterizable. That is, to sample from $q_{\theta}(z\g \varepsilon)$, we can first draw an auxiliary variable $u$ and then set $z$ as a deterministic function $h_{\theta}(\cdot)$ of the sampled $u$,
\begin{equation}\label{eq:implicit_rep}
	u\sim q(u), \;\; z=h_{\theta}(u\prm \varepsilon) \quad\equiv\quad z\sim q_{\theta}(z\g \varepsilon).
\end{equation}
The transformation $h_{\theta}(u\prm \varepsilon)$ is parameterized by the random variable $\varepsilon$ and the variational parameters $\theta$, but the auxiliary distribution $q(u)$ has no parameters. \Cref{fig:semi_implicit_illustration} illustrates this construction of the variational distribution.

The second assumption on the conditional $q_{\theta}(z\g \varepsilon)$ is that it is possible to evaluate the log-density $\log q_{\theta}(z\g \varepsilon)$ and its gradient with respect to $z$, $\nabla_{z} \log q_{\theta}(z\g \varepsilon)$. This is not a strong assumption; it holds for many reparameterizable distributions (e.g., Gaussian, Laplace, exponential, etc.).


\gls{USIVI} makes use of these two properties of the conditional $q_{\theta}(z\g \varepsilon)$ to derive unbiased estimates of the gradient of the \gls{ELBO} (see \Cref{sec:usivi}).

\parhead{Example: Gaussian conditional.}
As a simple example, consider a multivariate Gaussian distribution for the conditional $q_{\theta}(z\g \varepsilon)$. The parameters of the Gaussian are its mean $\mu_{\theta}(\varepsilon)$ and covariance $\Sigma_{\theta}(\varepsilon)$. Both parameters are given by neural networks with parameters $\theta$ and input $\varepsilon$.

The Gaussian meets the two assumptions outlined above. It is reparameterizable because it is in the location-scale family; the sampling process
\begin{equation*}
	\begin{split}
		& u\sim q(u)=\Ncal(u\g 0, I), \\
		& z=h_{\theta}(u\prm \varepsilon) =  \mu_{\theta}(\varepsilon) + \Sigma_{\theta}(\varepsilon)^{1/2} u
	\end{split}
\end{equation*}
generates a sample $z\sim q_{\theta}(z\g \varepsilon)$.

Furthermore, the Gaussian density and the gradient of the log-density can be evaluated. The latter is
\begin{equation*}
	\nabla_{z} \log q_{\theta}(z\g \varepsilon) = - \Sigma_{\theta}(\varepsilon)^{-1}\left(z - \mu_{\theta}(\varepsilon) \right).
\end{equation*}



\subsection{Unbiased Gradient Estimator}
\label{sec:usivi}

Here we derive the unbiased gradient estimators of the \gls{ELBO}. First, \gls{USIVI} uses the reparameterization $z=h_{\theta}(u\prm \varepsilon)$ (\Cref{eq:implicit_rep}) to rewrite the expectation in \Cref{eq:elbo} as an expectation with respect to $q(\varepsilon)$ and $q(u)$,
\begin{equation*}
	\Lcal(\theta) = \E{q(\varepsilon)q(u)}{\log p(x,z) - \log q_{\theta}(z)\Big|_{z=h_{\theta}(u\prm \varepsilon)}}.
\end{equation*}
To obtain the gradient of the \gls{ELBO} with respect to $\theta$, the gradient operator can now be pushed inside the expectation, as in the standard reparameterization method \citep{Kingma2014,Titsias2014_doubly,Rezende2014}. This gives two terms: one corresponding to the model and one corresponding to the entropy,
\begin{equation}\label{eq:grad_split}
	\nabla_{\theta}\Lcal(\theta) = \E{q(\varepsilon)q(u)}{g_{\theta}^{\textrm{mod}}(\varepsilon,u)+g_{\theta}^{\textrm{ent}}(\varepsilon,u)}.
\end{equation}
The term corresponding to the model is
\begin{equation}\label{eq:g_model}
	g_{\theta}^{\textrm{mod}}(\varepsilon,u) \triangleq \nabla_z \log p(x,z) \Big|_{z=h_{\theta}(u\prm \varepsilon)} \nabla_{\theta}h_{\theta}(u\prm \varepsilon);
\end{equation}
similarly, the term corresponding to the entropy is
\begin{equation}\label{eq:g_entropy}
	g_{\theta}^{\textrm{ent}}(\varepsilon,u) \triangleq - \nabla_z \log q_{\theta}(z) \Big|_{z=h_{\theta}(u\prm \varepsilon)} \nabla_{\theta}h_{\theta}(u\prm \varepsilon).
\end{equation}
To obtain this decomposition, we have applied the identity that the expected value of the score function is zero, $\E{q_{\theta}(z)}{\nabla_{\theta} \log q_{\theta}(z)} = 0$, which reduces the variance of the estimator \citep{Roeder2017}.

\gls{USIVI} estimates the model component in \Cref{eq:g_model} using samples from $q(\varepsilon)$ and $q(u)$. However, estimating the entropy component in \Cref{eq:g_entropy} is harder because the term $\nabla_{z} \log q_{\theta}(z)$ cannot be evaluated---the variational distribution $q_{\theta}(z)$ is an implicit distribution.

\gls{USIVI} addresses this issue rewriting \Cref{eq:g_entropy} as an expectation, therefore enabling Monte Carlo estimates of the entropy component of the gradient. In particular, \gls{USIVI} rewrites as an expectation the intractable log-density gradient in \Cref{eq:g_entropy},
\begin{equation}\label{eq:gradient_z_logq}
	\nabla_z \log q_{\theta}(z) = \E{q_{\theta}(\varepsilon\g z)}{\nabla_{z} \log q_{\theta}(z\g \varepsilon)}.
\end{equation}
We prove \Cref{eq:gradient_z_logq} below. This equation shows that the problematic gradient $\nabla_z \log q_{\theta}(z)$ can be expressed in terms of an expression that can be evaluated---the gradient of the log-conditional $\nabla_{z} \log q_{\theta}(z\g \varepsilon)$ can be evaluated by assumption (see \Cref{sec:sivi}). \gls{USIVI} rewrites the entropy term in \Cref{eq:g_entropy} using \Cref{eq:gradient_z_logq},
\begin{align}
	g_{\theta}^{\textrm{ent}}(\varepsilon,u) = & - \E{q_{\theta}(\varepsilon^\prime\g z)}{\nabla_{z} \log q_{\theta}(z\g \varepsilon^\prime)}  \Big|_{z=h_{\theta}(u\prm \varepsilon)}  \nonumber \\
	& \times \nabla_{\theta}h_{\theta}(u\prm \varepsilon). \label{eq:g_entropy_rewritten}
\end{align}
(We use the notation $\varepsilon^\prime$ to make it explicit that this variable is different from $\varepsilon$.)

The expectation in \Cref{eq:gradient_z_logq,eq:g_entropy_rewritten} is taken with respect to the distribution $q_{\theta}(\varepsilon\g z)\propto q_{\theta}(z\g \varepsilon)q(\varepsilon)$. We call this distribution the \emph{reverse conditional}. Although the conditional $q_{\theta}(z\g \varepsilon)$ has a simple form (by assumption, it is a reparameterizable distribution for which we can evaluate the density and its gradient), the reverse conditional is complex because the conditional $q_{\theta}(z\g \varepsilon)$ is parameterized by deep neural networks that take $\varepsilon$ as input. We show in \Cref{sec:algorithm} how to efficiently draw samples from the reverse conditional to obtain an estimator of the entropy component in \Cref{eq:g_entropy_rewritten}. (The main idea is to reuse a sample from the reverse conditional to initialize a sampler.)

We now prove \Cref{eq:gradient_z_logq} and then we show two examples that particularize these expressions for two choices of the conditionals: a multivariate Gaussian and a more general exponential family distribution.


\parhead{Proof of \Cref{eq:gradient_z_logq}.}
We show how to express the gradient $\nabla_z \log q_{\theta}(z)$ as an expectation. We start with the log-derivative identity,
\begin{equation}\nonumber
	\nabla_z \log q_{\theta}(z) = \frac{1}{q_{\theta}(z)}\nabla_z  q_{\theta}(z).
\end{equation}
Next we use the definition of the semi-implicit distribution $q_{\theta}(z)$ through a mixing distribution (\Cref{eq:implicit_dist_int}) and we push the gradient into the integral,
\begin{equation}\nonumber
	\begin{split}
		\nabla_z \log q_{\theta}(z) & = \frac{1}{q_{\theta}(z)}\nabla_z \int q_{\theta}(z\g \varepsilon)q(\varepsilon) d\varepsilon \\
		& = \frac{1}{q_{\theta}(z)}\int \nabla_z q_{\theta}(z\g \varepsilon)q(\varepsilon) d\varepsilon.
	\end{split}
\end{equation}
We now apply the log-derivative identity on the conditional $q_{\theta}(z\g \varepsilon)$,
\begin{equation}\nonumber
	\begin{split}
		\nabla_z \log q_{\theta}(z) = \frac{1}{q_{\theta}(z)}\int q_{\theta}(z\g \varepsilon) q(\varepsilon) \nabla_z \log q_{\theta}(z\g \varepsilon) d\varepsilon.
	\end{split}
\end{equation}
Finally, we apply Bayes' theorem to obtain \Cref{eq:gradient_z_logq}.\hfill\rlap{\hspace*{2em}$\square$}


\parhead{Example: Gaussian conditional.}
Consider the multivariate Gaussian example from \Cref{sec:sivi}. Substituting the gradient of the Gaussian log-density into \Cref{eq:g_entropy_rewritten}, we can write the entropy component of the gradient as
\begin{align*}
	g_{\theta}^{\textrm{ent}}(\varepsilon,u) = & \; \E{q_{\theta}(\varepsilon^\prime\g z)}{\Sigma_{\theta}(\varepsilon^\prime)^{-1}\left(z - \mu_{\theta}(\varepsilon^\prime) \right)}  \Big|_{z=h_{\theta}(u\prm \varepsilon)} \\
	& \times \nabla_{\theta}h_{\theta}(u\prm \varepsilon).
\end{align*}

\parhead{Example: Exponential family conditional.}
Now consider the more general example of a reparameterizable exponential family conditional distribution $q_{\theta}(z\g \varepsilon)$ with sufficient statistics $t(z)$ and natural parameter\footnote{We ignore the base measure in this definition; if needed it can be absorbed into $t(z)$.} $\eta_{\theta}(\varepsilon)$,
\begin{equation}
	q_{\theta}(z\g \varepsilon) \propto \exp\{ t(z)^\top\eta_{\theta}(\varepsilon) \}.
\end{equation}
Substituting the gradient $\nabla_z \log q_{\theta}(z\g \varepsilon)$ into \Cref{eq:g_entropy_rewritten}, we can obtain the entropy component of the gradient for a general (reparameterizable) exponential family distribution,
\begin{align*}
	g_{\theta}^{\textrm{ent}}(\varepsilon,u) = & \; - \nabla_z t(z)^\top \E{q_{\theta}(\varepsilon^\prime\g z)}{ \eta_{\theta}(\varepsilon^\prime) }  \Big|_{z=h_{\theta}(u\prm \varepsilon)} \\
	& \; \times \nabla_{\theta}h_{\theta}(u\prm \varepsilon).
\end{align*}


\subsection{Full Algorithm}
\label{sec:algorithm}

\gls{USIVI} estimates the gradient of the \gls{ELBO} using \Cref{eq:grad_split}, which decomposes the gradient as the expectation of the sum of the model component and the entropy component. \gls{USIVI} estimates the expectation using $S$ samples from $q(\varepsilon)$ and $q(u)$ ($S=1$ in practice); that is,
\begin{align*}
	& \nabla_{\theta}\Lcal(\theta)\approx \frac{1}{S} \sum_{s=1}^{S} \left(g_{\theta}^{\textrm{mod}}(\varepsilon_s,u_s) + g_{\theta}^{\textrm{ent}}(\varepsilon_s,u_s)\right), \\
	& \varepsilon_s \sim q(\varepsilon),\quad
	u_s \sim q(u).
\end{align*}
The model component is given in \Cref{eq:g_model} and the entropy component is given in \Cref{eq:g_entropy_rewritten}. While the model component can be evaluated (the gradients involved can be obtained using autodifferentiation tools), the entropy component is more challenging because \Cref{eq:g_entropy_rewritten} contains an expectation with respect to the reverse conditional $q_{\theta}(\varepsilon\g z)$. As this expectation is intractable, \gls{USIVI} forms a Monte Carlo estimator using samples $\varepsilon_s^\prime$ from the reverse conditional.

The reverse conditional is a complex distribution due to the complex dependency of the (direct) conditional $q_{\theta}(z\g \varepsilon)$ on the random variable $\varepsilon$. Consequently, sampling from the reverse conditional may be challenging.

\gls{USIVI} exploits the fact that the samples $\varepsilon_s$ that generated $z_s$ are also samples from the reverse conditional. This is because the sampling procedure in \Cref{eq:implicit_dist_samp} implies that each pair of samples $(z_s,\varepsilon_s)$ comes from the joint $q_{\theta}(z,\varepsilon)$, and thus $\varepsilon_s$ can be seen as a draw from the reverse conditional. 

Although $\varepsilon_s$ is a valid sample from the reverse conditional $q_{\theta}(\varepsilon \g z_s)$, setting $\varepsilon^\prime_s=\varepsilon_s$ in the estimation of the entropy component (\Cref{eq:g_entropy_rewritten}) would break the assumption that $\varepsilon_s^\prime$ and $\varepsilon_s$ are independent. Instead, \gls{USIVI} runs a \gls{MCMC} method, such as \gls{HMC} \citep{Neal2011}, to draw samples from the reverse conditional.\footnote{Note that this sampling algorithm does not require to evaluate the model $p(x,z)$, because the target distribution is $q_{\theta}(\varepsilon\g z_s)$.}
Crucially, \gls{USIVI} initializes the \gls{MCMC} chain at $\varepsilon_s$. In this way, there is no burn-in period in the \gls{MCMC} procedure, in the sense that the sampler starts from stationarity 
so that any subsequent \gls{MCMC} draw gives a sample from the reverse conditional \citep{Robert2005}. To reduce the correlation between the sample $\varepsilon_s^\prime$ and the initialization value $\varepsilon_s$, \gls{USIVI} runs more than one \gls{MCMC} iterations and allows for a short burn-in period. (In the experiments of \Cref{sec:experiments}, we use $10$ \gls{MCMC} iterations where only the final $5$ samples are used to form the Monte Carlo estimate.)

\gls{USIVI} then forms an unbiased estimator of the entropy component (\Cref{eq:g_entropy_rewritten}) using these samples from $q_{\theta}(\varepsilon \g z_s)$,
\begin{align*}
	& g_{\theta}^{\textrm{ent}}(\varepsilon_s,u_s) \approx - \nabla_{z} \log q_{\theta}(z\g \varepsilon^\prime_s) \nabla_{\theta}h_{\theta}(u_s\prm \varepsilon_s),\\
	& \varepsilon_s^\prime \sim q_{\theta}(\varepsilon\g z_s),\qquad z_s=h_{\theta}(u_s\prm \varepsilon_s).
\end{align*}
The full \gls{USIVI} algorithm is summarized in \Cref{alg:usivi}. For simplicity, in the description of the algorithm we assume one sample $\varepsilon_s^{\prime}$ for each sample $\varepsilon_s$; in practice we approximate each internal expectation under $q_{\theta}(\varepsilon^\prime \g z_s)$ with a few samples, i.e., the final $5$ samples from each $10$-length \gls{MCMC} run as mentioned above.
Code is publicly available in the authors' website.\footnote{See \url{https://github.com/franrruiz/uivi} for the code.}

\begin{algorithm}[tb]
	\caption{\Acrlong{USIVI}}
	\label{alg:usivi}
	\begin{algorithmic}
		\STATE {\bfseries Input:} data $x$, semi-implicit variational family $q_{\theta}(z)$
		\STATE {\bfseries Output:} variational parameters $\theta$
		\STATE Initialize $\theta$ randomly
		\FOR{iteration $t=1,2,\ldots,$}
			\STATE \verb|# Sample from q:|
			\STATE Sample $u_s\sim q(u)$ and $\varepsilon_s\sim q(\varepsilon)$
			\STATE Set $z_s = h_{\theta}(u_s\prm \varepsilon_s)$
			\STATE \verb|# Sample from reverse conditional:|
			\STATE Sample $\varepsilon_s^\prime \sim q_{\theta}(\varepsilon\g z_s)$ (\acrshort{HMC} initialized at $\varepsilon_s$)
			\STATE \verb|# Estimate the gradient:|
			\STATE Compute $g_{\theta}^{\textrm{mod}}(\varepsilon_s,u_s)$ (\Cref{eq:g_model})
			\STATE Compute $g_{\theta}^{\textrm{ent}}(\varepsilon_s,u_s)$ (\Cref{eq:g_entropy_rewritten}, approximate using $\varepsilon_s^\prime$)
			\STATE Compute $\widehat{\nabla}_{\theta}\Lcal = g_{\theta}^{\textrm{mod}}(\varepsilon_s,u_s) + g_{\theta}^{\textrm{ent}}(\varepsilon_s,u_s)$
			\STATE \verb|# Take gradient step:|
			\STATE Set $\theta \leftarrow \theta + \rho \cdot \widehat{\nabla}_{\theta}\Lcal$
		\ENDFOR
	\end{algorithmic}
\end{algorithm}

\section{RELATED WORK}
\label{sec:related}
\glsresetall

Among the methods to address the limitations of mean-field \gls{VI}, we can find methods that improve the mean-field posterior approximation using linear response estimates \citep{Giordano2015,Giordano2017}, or methods that add dependencies among the latent variables using a structured variational family \citep{Saul1996}, typically tailored to particular models \citep{Ghahramani1997,Titsias2011}. Other ways to add dependencies among the latent variables are mixtures \citep{Bishop1998,Gershman2012_nonparametric,Salimans2013,Guo2016,Miller2017}, copulas \citep{Tran2015,Han2016}, hierarchical models \citep{Ranganath2016,Tran2016,Maaloe2016}, or general invertible transformations of random variables \citep{Rezende2014,Kingma2014,Titsias2014_doubly,Kucukelbir2015,Kucukelbir2017}, including normalizing flows \citep{Rezende2015,Kingma2016,Papamakarios2017,Tomczak2016,Tomczak2017,Dinh2017}.
Other approaches use spectral methods \citep{Shi2018spectral} or define the variational distribution using sampling mechanisms \citep{Salimans2015,Maddison2017,Naesseth2017,Naesseth2018,Le2018,Grover2018}.

Implicit distributions develop a flexible variational family using non-invertible mappings parameterized by deep neural networks \citep{Mohamed2016,Nowozin2016,Huszar2017,Tran2017,Li2018,Mescheder2017,Shi2018}. The main issue of implicit distributions is density ratio estimation, which is often addressed using adversarial networks \citep{Goodfellow2014}. However, density ratio estimation becomes particularly difficult in high-dimensional settings \citep{Sugiyama2012}.

The method that is more closely related to ours is \gls{SIVI} \citep{Yin2018}. \gls{SIVI} combines a simple reparameterizable distribution with an implicit one to obtain a flexible variational family. To find the variational parameters, \gls{SIVI} maximizes a lower bound of the \gls{ELBO},
\begin{align}
	& \!\!\!\! \Lcal_{\textrm{SIVI}}^{(L)}(\theta) \! = \! \mathbb{E}_{\varepsilon\sim q(\varepsilon)} \!\Bigg[ \!\mathbb{E}_{z\sim q_{\theta}(z\g \varepsilon)}\!\Bigg[ \!\mathbb{E}_{\varepsilon^{(1)},\ldots,\varepsilon^{(L)}\sim q(\varepsilon)} \!\Bigg[ \!\log p(x,z) \nonumber \\
	& -\log \!\left(\! \frac{1}{L+1}\left( q_{\theta}(z\g \varepsilon) + \sum_{\ell=1}^L q_{\theta}(z\g \varepsilon^{(\ell)}) \right) \right) \!\Bigg]\Bigg]\Bigg], \label{eq:sivi_objective}
\end{align}
where $\Lcal_{\textrm{SIVI}}^{(L)}(\theta)\leq \Lcal(\theta)$ for any $\theta$.
At each iteration of the inference algorithm, the parameter $L$ must form a non-decreasing sequence. As the parameter $L$ grows to infinity, the lower bound $\Lcal_{\textrm{SIVI}}^{(L)}$ approaches the \gls{ELBO} in \Cref{eq:elbo}.
The intuition behind the \gls{SIVI} objective is to approximate the intractable marginalization $q_{\theta}(z) = \int q_{\theta}(z\g \varepsilon) q(\varepsilon) d\varepsilon$, which appears in the entropy component of the \gls{ELBO}, with $L+1$ draws from $q(\varepsilon)$.

\citet{Molchanov2019} have recently extended \gls{SIVI} in the context of deep generative models. They use a semi-implicit construction of both the variational distribution and the deep generative model that defines the prior. This results in a doubly semi-implicit architecture that allows building a sandwich estimator of the \gls{ELBO}. Similarly to \gls{SIVI}, the objective of doubly semi-implicit variational inference is a bound on the \gls{ELBO} that becomes tight as $L\rightarrow\infty$.

Finally, note that despite its similar name, the method of \citet{Figurnov2018} addresses a different problem. Specifically, it tackles the case where $q_{\theta}(z)$ is not reparameterizable but has a tractable density, for example a gamma or a beta distribution. In contrast, we address the problem where the variational distribution $q_{\theta}(z)$ is implicit.

\section{EXPERIMENTS}
\label{sec:experiments}
\glsresetall

We now apply \gls{USIVI} to assess the goodness of the resulting variational approximation and the computational complexity of the algorithm. As a baseline, we compare against \gls{SIVI}, which has been shown to outperform other approaches like mean-field \gls{VI} and be on par with \gls{MCMC} methods \citep{Yin2018}.

First, in \Cref{sec:experiments_toy} we run toy experiments on simple two-dimensional distributions. Then, in \Cref{sec:experiments_logreg,sec:experiments_vae} we turn to more realistic models, including Bayesian multinomial logistic regression and the \gls{VAE} \citep{Kingma2014}.

\subsection{Toy Experiments}
\label{sec:experiments_toy}

To showcase \gls{USIVI}, we approximate three synthetic distributions defined on a two-dimensional space: a banana distribution, a multimodal Gaussian, and an x-shaped Gaussian. Their densities are given in \Cref{tab:toy_distributions}.

\parhead{Variational family.}
To define the variational distribution, we choose a standard $3$-dimensional Gaussian prior for $q(\varepsilon)$. We use a Gaussian conditional $q_{\theta}(z\g \varepsilon)=\Ncal(z\g \mu_{\theta}(\varepsilon), \diag{\sigma})$, whose mean is parameterized by a neural network with two hidden layers of $50$ ReLu units each. We set a diagonal covariance that we also optimize (for simplicity, the covariance does not depend on $\varepsilon$). Thus, the variational parameters are the neural network weights and intercepts ($\theta$) and the variances ($\sigma$).

\parhead{Experimental setup.}
We run $50{,}000$ iterations of \Cref{alg:usivi}. We run $10$ \gls{HMC} iterations to draw samples from the reverse conditional $q_{\theta}(\varepsilon\g z)$ ($5$ for burn-in and $5$ actual samples), with $5$ leapfrog steps \citep{Neal2011}. We set the stepsize using RMSProp 
\citep{Tieleman2012}; at each iteration $t$ we set $\rho^{(t)}=\eta /(1+\sqrt{G^{(t)}})$, where $\eta$ is the learning rate, and the updates of $G^{(t)}$ depend on the gradient $\widehat{\nabla}_{\theta} \Lcal^{(t)}$ as $G^{(t)}=0.9G^{(t-1)} + 0.1 (\widehat{\nabla}_{\theta} \Lcal^{(t)})^2$. We set the learning rate $\eta=0.01$ for the network parameters and $\eta=0.002$ for the covariance, and we additionally decrease the learning rate by a factor of $0.9$ every $3{,}000$ iterations.

\begin{figure*}[t]
	\centering
	\begin{tabular}{ccc}
		\includegraphics[width=0.26\textwidth]{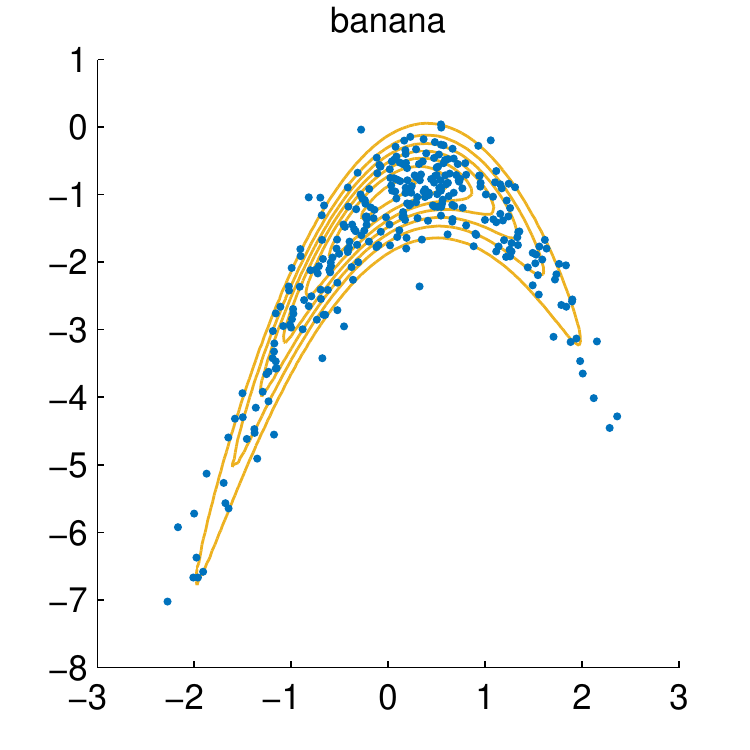} &
		\includegraphics[width=0.26\textwidth]{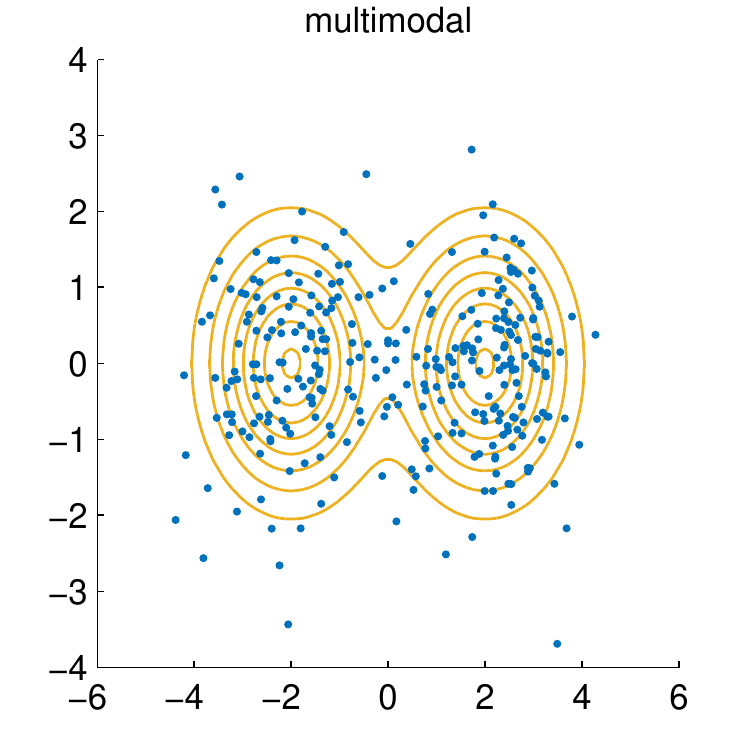} &
		\includegraphics[width=0.26\textwidth]{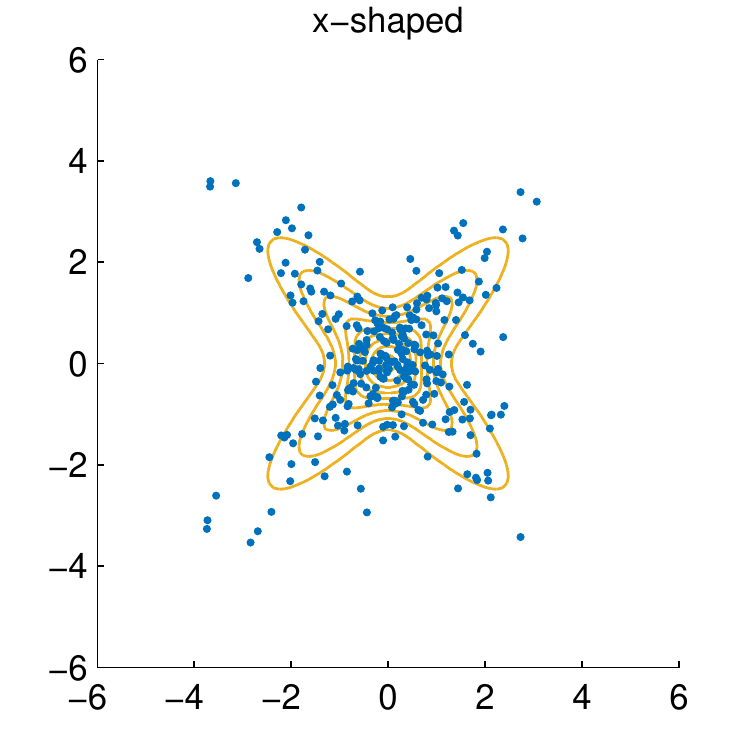} 
	\end{tabular}
	\vspace*{-10pt}
	\caption{The samples from the variational distribution fitted with \gls{USIVI} (blue) match the shape of the true synthetic target distributions (orange) considered in \Cref{sec:experiments_toy}.\label{fig:experiments_toy}}
	\vspace*{-10pt}
\end{figure*}

\parhead{Results.} \Cref{fig:experiments_toy} shows the contour plot of the synthetic distributions, together with $300$ samples from the fitted variational distribution. \gls{USIVI} produces samples that match well the shape of the target distributions.


\begin{table}[t]
	\vspace*{5pt}
	\centering
 	\small
	{\setlength{\tabcolsep}{1.2pt}
	\begin{tabular}{cc} \toprule
		name & $p(z)$ \\ \midrule
		banana & {\scriptsize
				$ \Ncal\left( \left[\begin{array}{c} z_1 \\ z_2+z_1^2+1 \end{array}\right] \bigg| \left[\begin{array}{c} 0 \\ 0 \end{array}\right], \left[\begin{array}{cc} 1 & 0.9 \\ 0.9 & 1 \end{array}\right] \right)$} \vspace*{5pt} \\ 
		multimodal & {\scriptsize $0.5\Ncal\left(z\;\bigg| \left[\begin{array}{c} -2 \\ 0 \end{array}\right], I\right) + 0.5\Ncal\left(z\;\bigg| \left[\begin{array}{c} 2 \\ 0 \end{array}\right], I\right)$ } \vspace*{5pt} \\
		x-shaped & {\scriptsize $0.5\Ncal\!\left(z\;\bigg|\; 0, \left[\! \arraycolsep=1.2pt \begin{array}{cc} 2 & 1.8  \\ 1.8 & 2 \end{array} \!\right]\right)\! + 0.5\Ncal\!\left(z\;\bigg| \; 0, \left[\! \arraycolsep=0.9pt \begin{array}{cc} 2 & -1.8 \\ -1.8 & 2 \end{array}\!\right]\right)$ } \vspace*{1pt} \\ \bottomrule
	\end{tabular} }
	\vspace*{-4pt}
	\caption{Synthetic distributions used in the toy experiment.\label{tab:toy_distributions}}
	\vspace*{-5pt}
\end{table}

\subsection{Bayesian Multinomial Logistic Regression}
\label{sec:experiments_logreg}

We now consider Bayesian multinomial logistic regression. For a dataset of $N$ features $x_n$ and labels $y_n\in\{1,\ldots,K\}$, the model is $p(z)\prod_n p(y_n\g x_n, z)$, where $z$ denotes the latent weights and biases. We set the prior $p(z)$ to be Gaussian with identity covariance and zero mean; the categorical likelihood is $p(y_n=k\g x_n, z)\propto\exp(x_n^\top z_k + z_{0k})$.


\parhead{Datasets.} We use two datasets, \textsc{mnist}\footnote{\url{http://yann.lecun.com/exdb/mnist}}
and \textsc{hapt},\footnote{\url{https://archive.ics.uci.edu/ml/datasets/Smartphone-Based+Recognition+of+Human+Activities+and+Postural+Transitions}}
both available online.
\textsc{mnist} contains $60{,}000$ training and $10{,}000$ test instances of $28\times 28$ images of hand-written digits; thus there are $K=10$ classes. We divide pixel values by $255$ so that each feature is bounded between $0$ and $1$.
\textsc{hapt}
\citep{ReyesOrtiz2016} is a human activity recognition dataset. It contains $7{,}767$ training and $3{,}162$ test $561$-dimensional measurements captured by the sensors on a smartphone. There are $K=12$ activities, including static postures (e.g., standing), dynamic activities (e.g., walking), and postural transitions (e.g., stand-to-sit).

\parhead{Variational family.} We use the variational family described in \Cref{sec:experiments_toy}, namely, a Gaussian prior $q(\varepsilon)$ and Gaussian conditional $q_{\theta}(z\g \varepsilon)$ with diagonal covariance. We set the dimensionality of $\varepsilon$ to $100$, and we use $200$ hidden units on each of the two hidden layers of the neural network that parameterizes the mean of the Gaussian conditional.

\parhead{Experimental setup.} We run $100{,}000$ iterations of \gls{USIVI}, with the same experimental setup described in \Cref{sec:experiments_toy}.
To speed up the procedure, we subsample minibatches of data at each iteration of the algorithm \citep{Hoffman2013}. We use a minibatch size of $2{,}000$ for \textsc{mnist} and $863$ for \textsc{hapt}.
For the comparison with \gls{SIVI}, we set the parameter $L=200$ in \Cref{eq:sivi_objective}. We use the same initialization of the variational parameters for both \gls{SIVI} and \gls{USIVI}.

\begin{figure*}[tb]
	\centering
	\begin{tabular}{cc}
		\includegraphics[width=0.48\textwidth]{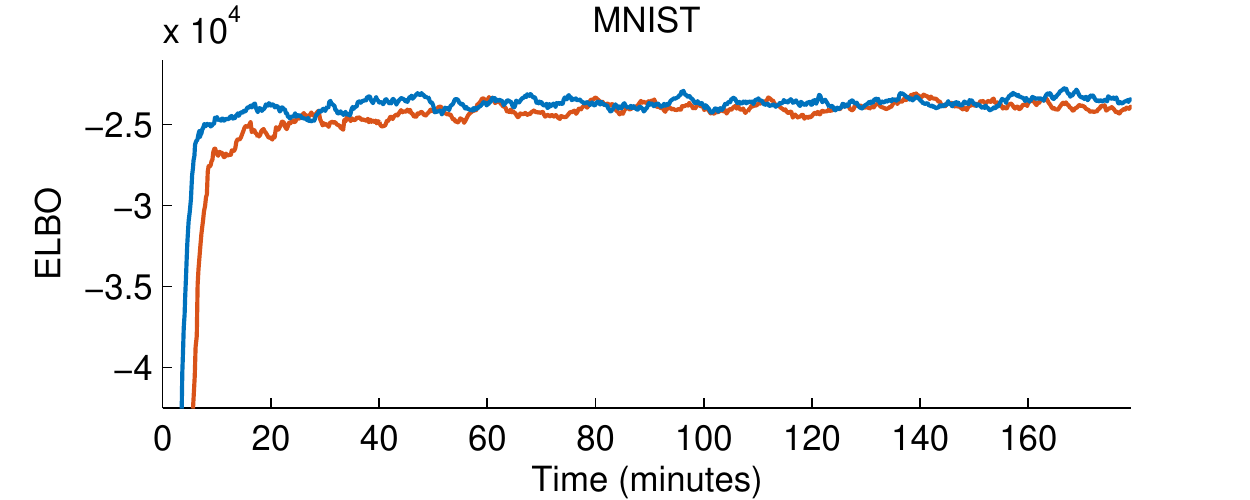} & 
		\includegraphics[width=0.48\textwidth]{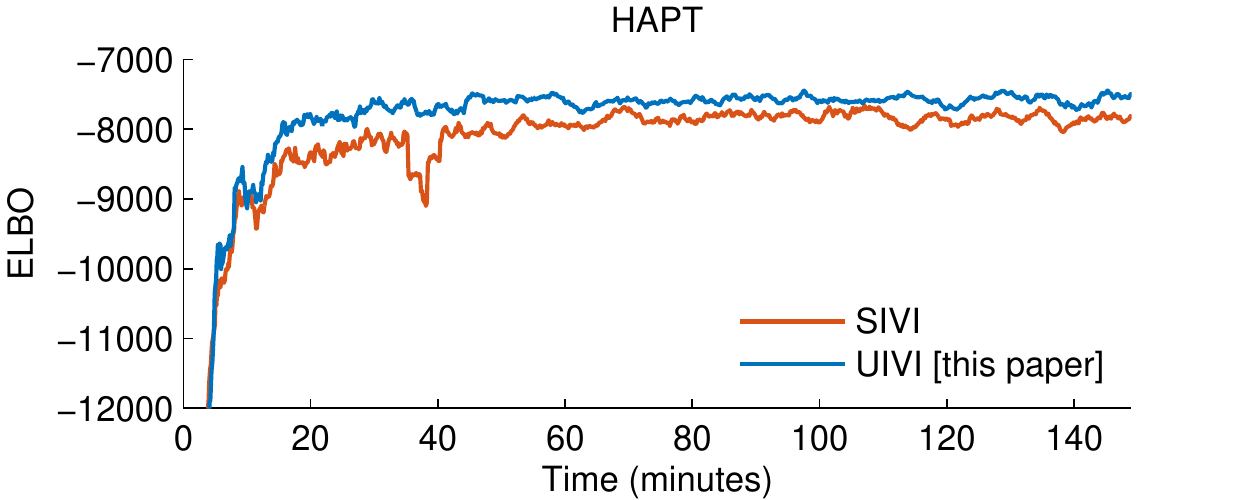} \\
		\includegraphics[width=0.48\textwidth]{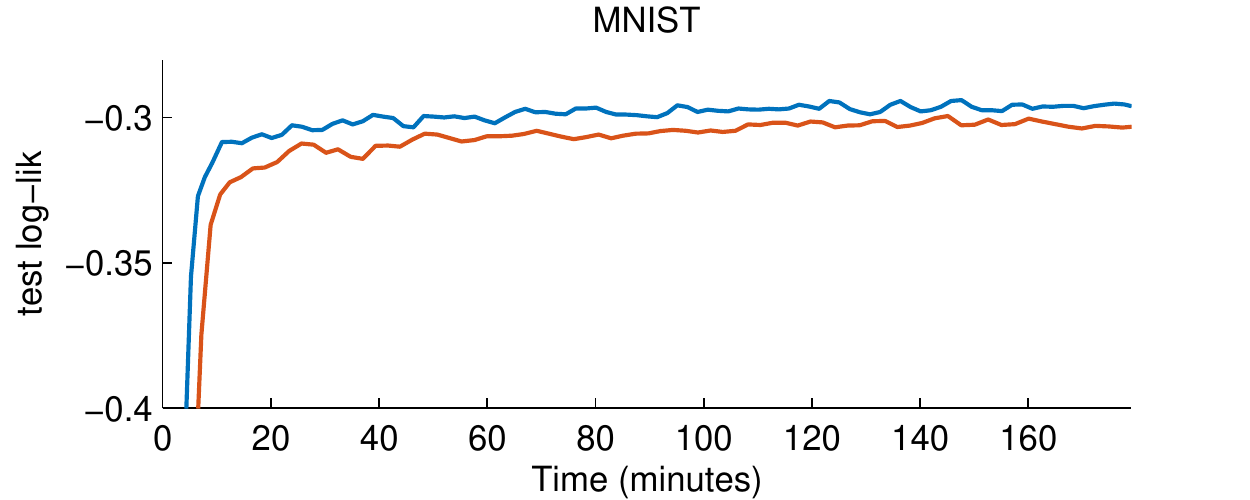} & 
		\includegraphics[width=0.48\textwidth]{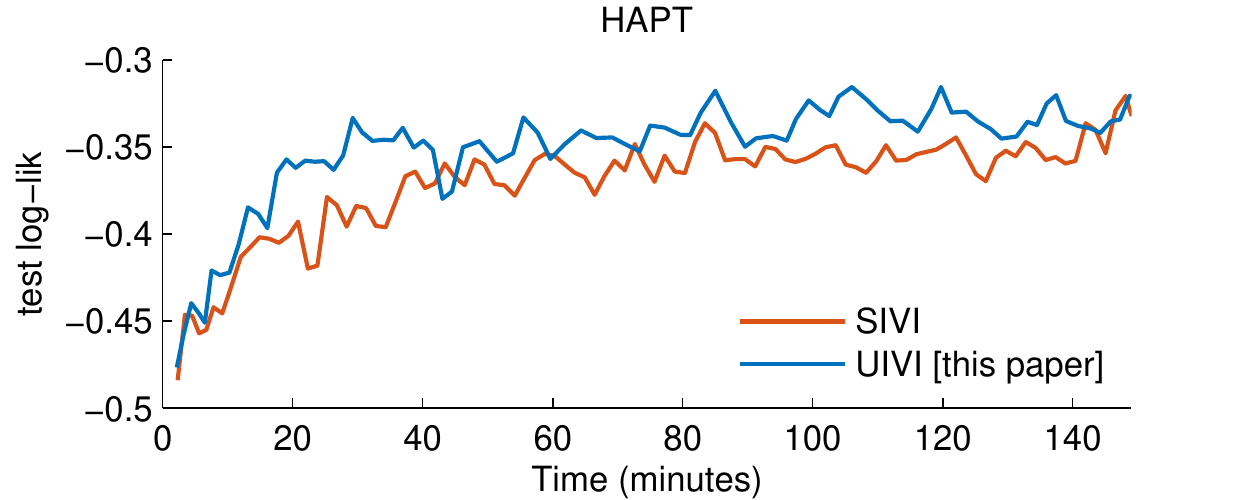}
	\end{tabular}
	\vspace*{-4pt}
	\caption{Estimates of the \acrshort{ELBO} and the test log-likelihood as a function of wall-clock time for the Bayesian multinomial logistic regression model (\Cref{sec:experiments_logreg}). Compared to \gls{SIVI} (red), \gls{USIVI} (blue) achieves a better bound on the marginal likelihood and has better predictive performance. \label{fig:experiments_softmaxclass}}
	\vspace*{-4pt}
\end{figure*}

\parhead{Results.} We found that the time per iteration was comparable for both methods. On average, \gls{SIVI} took $0.14$ seconds per iteration on \textsc{mnist} and $0.09$ seconds on \textsc{hapt}, while \gls{USIVI} took $0.11$ and $0.10$ seconds, respectively.

We obtain a Monte Carlo estimate of the \gls{ELBO} every $100$ iterations (we use $100$ samples, and we use $10{,}000$ samples from $q(\varepsilon)$ to approximate the intractable entropy term). \Cref{fig:experiments_softmaxclass} (top) shows the \gls{ELBO} estimates; the plot has been smoothed using a rolling window of size $20$ for easier visualization. \gls{USIVI} provides a similar bound on the marginal likelihood than \gls{SIVI} on \textsc{mnist} and a slightly tighter bound on \textsc{hapt}.

In addition, we also estimate the predictive log-likelihood on the test set every $1{,}000$ iterations (we use $8{,}000$ samples from the variational distribution to form this estimator). \Cref{fig:experiments_softmaxclass} (bottom) shows the test log-likelihood as a function of the wall-clock time for both methods and datasets; the plot has been smoothed with a rolling window of size $2$. \gls{USIVI} achieves better predictions on both datasets.

\begin{figure}[tb]
	\centering
	\includegraphics[width=0.48\textwidth]{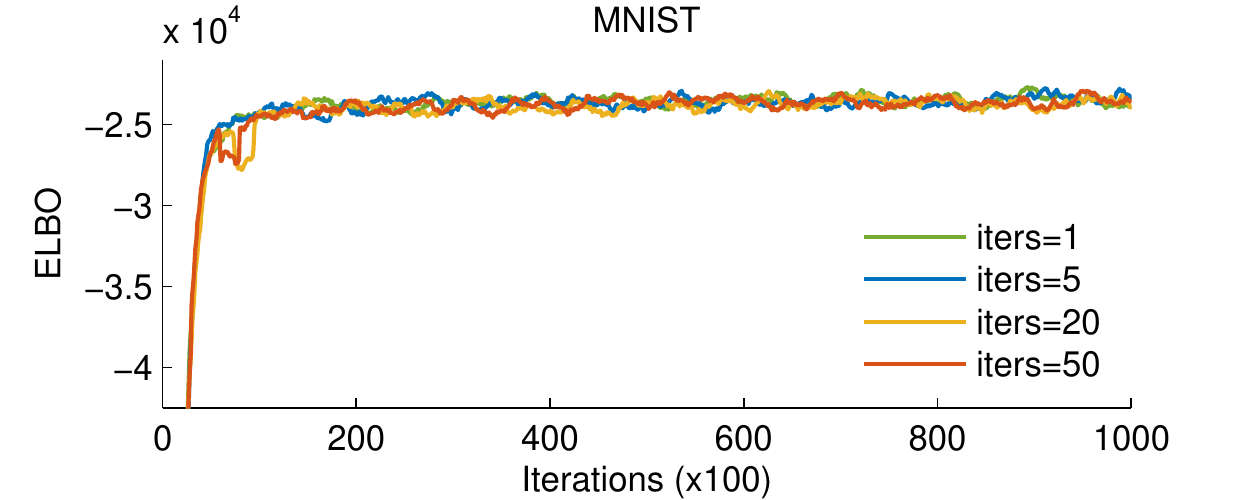}
	\caption{Estimates of the \acrshort{ELBO} for the Bayesian multinomial logistic regression model (\Cref{sec:experiments_logreg}), obtained with \acrshort{USIVI} under four different settings, which differ only in the number of \acrshort{HMC} iterations. The number of \acrshort{HMC} iterations in \acrshort{USIVI} has a small impact on the results.\label{fig:experiments_softmaxclass_varyIters}}
	\vspace*{-10pt}
\end{figure}

Finally, we study the impact of the number of \gls{HMC} iterations on the results. In \Cref{fig:experiments_softmaxclass_varyIters}, we plot the \gls{ELBO} as a function of the iterations of the variational algorithm in four different settings on the \textsc{mnist} dataset. Each of these settings corresponds to a different number of \gls{HMC} iterations for both burn-in and sampling periods, ranging from $1$ to $50$ iterations (the standard setting of \Cref{fig:experiments_softmaxclass} corresponds to $5$ iterations). We conclude that the number of \gls{HMC} iterations does not have a significant impact on the results. (Although not included here, the plot for the test log-likelihood shows no significant differences either.)

\subsection{Variational Autoencoders}
\label{sec:experiments_vae}

The \gls{VAE} \citep{Kingma2014} defines a conditional likelihood $p_{\phi}(x_n\g z_n)$ given the latent variable $z_n$,  parameterized by a neural network with parameters $\phi$. The goal is to learn the parameters $\phi$, for which the \gls{VAE} introduces an amortized variational distribution $q_{\theta}(z_n\g x_n)$. In the standard \gls{VAE}, the variational distribution is Gaussian; we use instead a semi-implicit variational distribution.

\parhead{Datasets.} We use two datasets: (i) the binarized \textsc{mnist} data \citep{Salakhutdinov2008_quantitative}, which contains $50{,}000$ training images and $10{,}000$ test images of handwritten digits; and (ii) the binarized Fashion-\textsc{mnist} data \citep{Xiao2017}, which contains $60{,}000$ training images and $10{,}000$ test images of clothing items. 
We binarize the Fashion-\textsc{mnist} images with a threshold at $0.5$. Images in both datasets are of size $28\times 28$ pixels.

\parhead{Variational family.} We use the variational family described in \Cref{sec:experiments_toy} with Gaussian prior and Gaussian conditional. Since the variational distribution is amortized, we let the conditional $q_{\theta}(z_n\g \varepsilon_n, x_n)$ depend on the observation $x_n$, such that the variational distribution is $q_{\theta}(z_n\g x_n) = \int q(\varepsilon_n) q_{\theta}(z_n\g \varepsilon_n, x_n)d\varepsilon_n$. We obtain the mean of the Gaussian conditional as the output of a neural network having as inputs both $x_n$ and $\varepsilon_n$. We set the dimensionality of $\varepsilon_n$ to $10$ and the width of each the two hidden layers of the neural network to $200$.

For comparisons, we also fit a standard \gls{VAE} \citep{Kingma2014}. The standard \gls{VAE} uses an explicit Gaussian distribution whose mean and covariance are functions of the input, i.e., $q_{\theta}(z_n\g x_n)=\Ncal(z_n\g\mu_{\theta}(x_n),\Sigma_{\theta}(x_n))$. The mean and covariance are parameterized using two separate neural networks with the same structure described above, and the covariance is set to be diagonal. The neural network for the covariance has softplus activations in the output layer, i.e., $\textrm{softplus}(x)=\log(1+e^x)$.

\parhead{Experimental setup.} For the generative model $p_{\phi}(x_n\g z_n)$ we use a factorized Bernoulli distribution. We use a two-hidden-layer neural network with $200$ hidden units on each hidden layer, whose sigmoidal outputs define the means of the Bernoulli 
 distribution. We set the prior $p(z_n)=\Ncal(z_n\g 0,I)$ and the dimensionality of $z_n$ to $10$.
We run $400{,}000$ iterations of each method (explicit variational distribution, \gls{SIVI}, and \gls{USIVI}), using the same initialization and a minibatch of size $100$. We set the \gls{SIVI} parameter $L=100$ so that both \gls{SIVI} and \gls{USIVI} have similar complexity (see below).
We set the learning rate $\eta=10^{-3}$ for the network parameters of the variational Gaussian conditional, $\eta=2\cdot 10^{-4}$ for its covariance (we also set $\eta=2\cdot 10^{-4}$ for the network that parameterizes the covariance of the explicit distribution), and $\eta=10^{-3}$ for the network parameters of the generative model. We reduce the learning rate by a factor of $0.9$ every $15{,}000$ iterations.

\begin{table}[t]
	\centering
 	\small
	\begin{tabular}{ccc} \toprule
		& \multicolumn{2}{c}{average test log-likelihood} \\
		method & \textsc{mnist} & Fashion-\textsc{mnist} \\ \midrule
		Explicit (standard \acrshort{VAE}) & $-98.29$ & $-126.73$ \\
		\acrshort{SIVI} & $-97.77$ & $-121.53$ \\
		\acrshort{USIVI} [this paper] & $\mathbf{-94.09}$ & $\mathbf{-110.72}$ \\ \bottomrule
	\end{tabular}
	\caption{Estimates of the marginal log-likelihood on the test set for the \gls{VAE} (\Cref{sec:experiments_vae}). \gls{USIVI} gives better predictive performance than \gls{SIVI}.\label{tab:experiments_vae_llh}}
\end{table}

\begin{figure}[t]
	\centering
	\begin{subfigure}[t]{0.5\textwidth}
		\centering
		\includegraphics[width=0.9\textwidth]{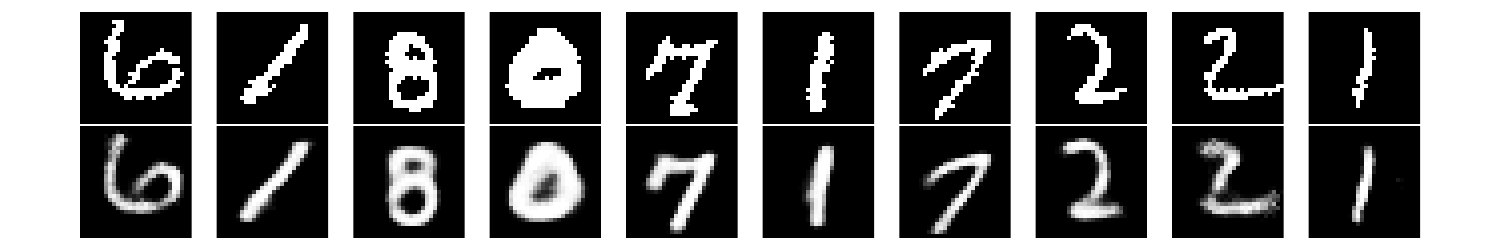}
		\vspace*{-4pt}
		\caption{\textsc{mnist} images.}
		\vspace*{4pt}
	\end{subfigure}
	\begin{subfigure}[t]{0.5\textwidth}
		\centering
		\includegraphics[width=0.9\textwidth]{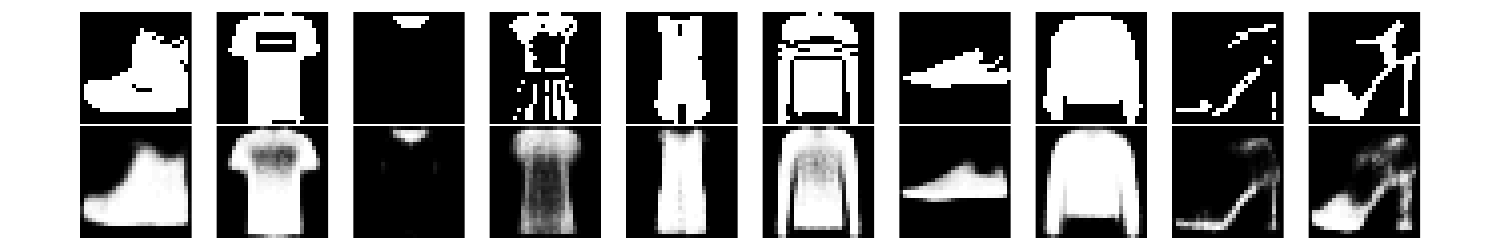}
		\vspace*{-2pt}
		\caption{Fashion-\textsc{mnist} images.}
	\end{subfigure}
	\caption{Ten images reconstructed with the \gls{VAE} model fitted with \gls{USIVI} (\Cref{sec:experiments_vae}). For each dataset, the top row shows training instances; the bottom row corresponds to the reconstructed images. \label{fig:experiments_vae_reconstruct}}
	\vspace*{-5pt}
\end{figure}

\parhead{Results.} We estimate the marginal likelihood on the test set using importance sampling,
\begin{equation*}
	\begin{split}
		& \log p(x_n) \approx \log \frac{1}{S} \sum_{s=1}^{S} \frac{p_{\phi}(x_n\g z_n^{(s)})p(z_n^{(s)})}{\frac{1}{M}\sum_{m=1}^M q_{\theta}(z_n^{(s)}\g \varepsilon_n^{(m)},x_n)},\\
		& z_n^{(s)} \sim q_{\theta}(z_n\g x_n),\quad \varepsilon_n^{(m)}\sim q(\varepsilon),
	\end{split}
\end{equation*}
where we set $S=1{,}000$ and $M=10{,}000$ samples.

\Cref{tab:experiments_vae_llh} shows the estimated values of the test marginal likelihood for all methods and datasets. \gls{USIVI} provides better predictive performance than \gls{SIVI}, which in turn gives better predictions than the explicit Gaussian approximation.

In terms of computational complexity, the average time per iteration is similar for \gls{USIVI} and \gls{SIVI}. On \textsc{mnist}, it is $0.14$ seconds for \gls{USIVI} and $0.16$ seconds for \gls{SIVI}; on Fashion-\textsc{mnist}, it is $0.13$ seconds for \gls{USIVI} and $0.17$ for \gls{SIVI}.



Finally, we show in \Cref{fig:experiments_vae_reconstruct} ten training images from each dataset, together with the corresponding images reconstructed using the \gls{VAE} fitted with \gls{USIVI}. We reconstruct an image by first sampling $z_n\sim q_{\theta}(z_n\g x_n)$ and then setting the reconstructed $\hat{x}_n$ to the mean given by the generative model $p_{\phi}(x_n\g z_n)$. We conclude that \gls{USIVI} is an effective method to optimize the \gls{VAE} model.

\section{CONCLUSION}
\label{sec:conclusion}
\glsresetall

We have developed \gls{USIVI}, a method to approximate a target distribution with an expressive variational distribution. The variational distribution is implicit, and it is obtained through a reparameterizable distribution whose parameters follow a flexible distribution, similarly to \gls{SIVI} \citep{Yin2018}. In contrast to \gls{SIVI}, \gls{USIVI} directly optimizes the \gls{ELBO} rather than a bound. For that, \gls{USIVI} expresses the gradient of the \gls{ELBO} as an expectation, enabling Monte Carlo estimates of the gradient. Compared to \gls{SIVI}, we show that \gls{USIVI} achieves better \gls{ELBO} and predictive performance for Bayesian multinomial logistic regression and \acrlong{VAE}.


\subsubsection*{Acknowledgements}
Francisco J.\ R.\ Ruiz is supported by the EU Horizon 2020 programme (Marie Sk\l{}odowska-Curie Individual Fellowship, grant agreement 706760).

\bibliography{fjrrLibrary}

\begin{thebibliography}{62}
\providecommand{\natexlab}[1]{#1}
\providecommand{\url}[1]{\texttt{#1}}
\expandafter\ifx\csname urlstyle\endcsname\relax
  \providecommand{\doi}[1]{doi: #1}\else
  \providecommand{\doi}{doi: \begingroup \urlstyle{rm}\Url}\fi

\bibitem[Bishop et~al.(1998)Bishop, Lawrence, Jaakkola, and Jordan]{Bishop1998}
Bishop, C.~M., Lawrence, N.~D., Jaakkola, T.~S., and Jordan, M.~I.
\newblock Approximating posterior distributions in belief networks using
  mixtures.
\newblock In \emph{Advances in Neural Information Processing Systems}, 1998.

\bibitem[Blei et~al.(2017)Blei, Kucukelbir, and McAuliffe]{Blei2017}
Blei, D.~M., Kucukelbir, A., and McAuliffe, J.~D.
\newblock Variational inference: {A} review for statisticians.
\newblock \emph{Journal of the American Statistical Association}, 112\penalty0
  (518):\penalty0 859--877, 2017.

\bibitem[Carbonetto et~al.(2009)Carbonetto, King, and Hamze]{Carbonetto2009}
Carbonetto, P., King, M., and Hamze, F.
\newblock A stochastic approximation method for inference in probabilistic
  graphical models.
\newblock In \emph{Advances in Neural Information Processing Systems}, 2009.

\bibitem[Dinh et~al.(2017)Dinh, Sohl-Dickstein, and Bengio]{Dinh2017}
Dinh, L., Sohl-Dickstein, J., and Bengio, S.
\newblock Density estimation using real {NVP}.
\newblock In \emph{International Conference on Learning Representations}, 2017.

\bibitem[Figurnov et~al.(2018)Figurnov, Mohamed, and Mnih]{Figurnov2018}
Figurnov, M., Mohamed, S., and Mnih, A.
\newblock Implicit reparameterization gradients.
\newblock In \emph{Advances in Neural Information Processing Systems}, 2018.

\bibitem[Gershman et~al.(2012)Gershman, Hoffman, and
  Blei]{Gershman2012_nonparametric}
Gershman, S.~J., Hoffman, M.~D., and Blei, D.~M.
\newblock Nonparametric variational inference.
\newblock In \emph{International Conference on Machine Learning}, 2012.

\bibitem[Ghahramani \& Beal(2001)Ghahramani and Beal]{Ghahramani2001}
Ghahramani, Z. and Beal, M.~J.
\newblock Propagation algorithms for variational {B}ayesian learning.
\newblock In \emph{Advances in Neural Information Processing Systems}, 2001.

\bibitem[Ghahramani \& Jordan(1997)Ghahramani and Jordan]{Ghahramani1997}
Ghahramani, Z. and Jordan, M.~I.
\newblock Factorial hidden {M}arkov models.
\newblock \emph{Machine Learning}, 29\penalty0 (2--3):\penalty0 245--273, 1997.

\bibitem[Giordano et~al.(2015)Giordano, Broderick, and Jordan]{Giordano2015}
Giordano, R.~J., Broderick, T., and Jordan, M.~I.
\newblock Linear response methods for accurate covariance estimates from mean
  field variational {B}ayes.
\newblock In \emph{Advances in Neural Information Processing Systems}, 2015.

\bibitem[Giordano et~al.(2017)Giordano, Broderick, and Jordan]{Giordano2017}
Giordano, R.~J., Broderick, T., and Jordan, M.~I.
\newblock Covariances, robustness, and variational {B}ayes.
\newblock In \emph{arXiv:1709.02536}, 2017.

\bibitem[Goodfellow et~al.(2014)Goodfellow, Pouget-Abadie, Mirza, Xu,
  Warde-Farley, Ozair, Courville, and Bengio]{Goodfellow2014}
Goodfellow, I., Pouget-Abadie, J., Mirza, M., Xu, B., Warde-Farley, D., Ozair,
  S., Courville, A., and Bengio, Y.
\newblock Generative adversarial nets.
\newblock In \emph{Advances in Neural Information Processing Systems}, 2014.

\bibitem[Grover et~al.(2018)Grover, Gummadi, L\'{a}zaro-Gredilla, Schuurmans,
  and Ermon]{Grover2018}
Grover, A., Gummadi, R., L\'{a}zaro-Gredilla, M., Schuurmans, D., and Ermon, S.
\newblock Variational rejection sampling.
\newblock In \emph{Artificial Intelligence and Statistics}, 2018.

\bibitem[Guo et~al.(2016)Guo, Wang, Fan, Broderick, and Dunson]{Guo2016}
Guo, F., Wang, X., Fan, K., Broderick, T., and Dunson, D.~B.
\newblock Boosting variational inference.
\newblock In \emph{arXiv:1611.05559}, 2016.

\bibitem[Han et~al.(2016)Han, Liao, Dunson, and Carin]{Han2016}
Han, S., Liao, X., Dunson, D.~B., and Carin, L.
\newblock Variational {G}aussian copula inference.
\newblock In \emph{Artificial Intelligence and Statistics}, 2016.

\bibitem[Hoffman et~al.(2013)Hoffman, Blei, Wang, and Paisley]{Hoffman2013}
Hoffman, M.~D., Blei, D.~M., Wang, C., and Paisley, J.
\newblock Stochastic variational inference.
\newblock \emph{Journal of Machine Learning Research}, 14:\penalty0 1303--1347,
  May 2013.

\bibitem[Husz\'{a}r(2017)]{Huszar2017}
Husz\'{a}r, F.
\newblock Variational inference using implicit distributions.
\newblock In \emph{arXiv:1702.08235}, 2017.

\bibitem[Jaakkola \& Jordan(1998)Jaakkola and Jordan]{Jaakkola1998}
Jaakkola, T.~S. and Jordan, M.~I.
\newblock Improving the mean field approximation via the use of mixture
  distributions.
\newblock In \emph{Learning in Graphical Models}, pp.\  163--173, 1998.

\bibitem[Jordan(1999)]{Jordan1999learn}
Jordan, M.~I. (ed.).
\newblock \emph{Learning in Graphical Models}.
\newblock MIT Press, Cambridge, MA, USA, 1999.

\bibitem[Kingma \& Welling(2014)Kingma and Welling]{Kingma2014}
Kingma, D.~P. and Welling, M.
\newblock Auto-encoding variational {B}ayes.
\newblock In \emph{International Conference on Learning Representations}, 2014.

\bibitem[Kingma et~al.(2016)Kingma, Salimans, Jozefowicz, Chen, Sutskever, and
  Welling]{Kingma2016}
Kingma, D.~P., Salimans, T., Jozefowicz, R., Chen, X., Sutskever, I., and
  Welling, M.
\newblock Improved variational inference with inverse autoregressive flow.
\newblock In \emph{Advances in Neural Information Processing Systems}, 2016.

\bibitem[Kucukelbir et~al.(2015)Kucukelbir, Ranganath, Gelman, and
  Blei]{Kucukelbir2015}
Kucukelbir, A., Ranganath, R., Gelman, A., and Blei, D.~M.
\newblock Automatic variational inference in {S}tan.
\newblock In \emph{Advances in Neural Information Processing Systems}, 2015.

\bibitem[Kucukelbir et~al.(2017)Kucukelbir, Tran, Ranganath, Gelman, and
  Blei]{Kucukelbir2017}
Kucukelbir, A., Tran, D., Ranganath, R., Gelman, A., and Blei, D.~M.
\newblock Automatic differentiation variational inference.
\newblock \emph{Journal of Machine Learning Research}, 18\penalty0
  (14):\penalty0 1--45, 2017.

\bibitem[Le et~al.(2018)Le, Igl, Rainforth, Jin, and Wood]{Le2018}
Le, T.~A., Igl, M., Rainforth, T., Jin, T., and Wood, F.
\newblock Auto-encoding sequential {M}onte-{C}arlo.
\newblock In \emph{International Conference on Learning Representations}, 2018.

\bibitem[Li \& Turner(2018)Li and Turner]{Li2018}
Li, Y. and Turner, R.~E.
\newblock Gradient estimators for implicit models.
\newblock In \emph{International Conference on Learning Representations}, 2018.

\bibitem[Maal{\o}e et~al.(2016)Maal{\o}e, S{\o}nderby, S{\o}nderby, and
  Winther]{Maaloe2016}
Maal{\o}e, L., S{\o}nderby, C.~K., S{\o}nderby, K., and Winther, O.
\newblock Auxiliary deep generative models.
\newblock In \emph{International Conference on Machine Learning}, 2016.

\bibitem[Maddison et~al.(2017)Maddison, Lawson, Tucker, Heess, Norouzi, Mnih,
  Doucet, and Teh]{Maddison2017}
Maddison, C.~J., Lawson, D., Tucker, G., Heess, N., Norouzi, M., Mnih, A.,
  Doucet, A., and Teh, Y.~W.
\newblock Filtering variational objectives.
\newblock In \emph{Advances in Neural Information Processing Systems}, 2017.

\bibitem[Mescheder et~al.(2017)Mescheder, Nowozin, and Geiger]{Mescheder2017}
Mescheder, L., Nowozin, S., and Geiger, A.
\newblock Adversarial variational {B}ayes: Unifying variational autoencoders
  and generative adversarial networks.
\newblock In \emph{International Conference on Machine Learning}, 2017.

\bibitem[Miller et~al.(2017)Miller, Foti, and Adams]{Miller2017}
Miller, A.~C., Foti, N., and Adams, R.~P.
\newblock Variational boosting: Iteratively refining posterior approximations.
\newblock In \emph{International Conference on Machine Learning}, 2017.

\bibitem[Mohamed \& Lakshminarayanan(2016)Mohamed and
  Lakshminarayanan]{Mohamed2016}
Mohamed, S. and Lakshminarayanan, B.
\newblock Learning in implicit generative models.
\newblock In \emph{arXiv:1610.03483}, 2016.

\bibitem[Molchanov et~al.(2019)Molchanov, Kharitonov, Sobolev, and
  Vetrov]{Molchanov2019}
Molchanov, D., Kharitonov, V., Sobolev, A., and Vetrov, D.
\newblock Doubly semi-implicit variational inference.
\newblock In \emph{Artificial Intelligence and Statistics}, 2019.

\bibitem[Naesseth et~al.(2017)Naesseth, Ruiz, Linderman, and
  Blei]{Naesseth2017}
Naesseth, C., Ruiz, F. J.~R., Linderman, S., and Blei, D.~M.
\newblock Reparameterization gradients through acceptance-rejection methods.
\newblock In \emph{Artificial Intelligence and Statistics}, 2017.

\bibitem[Naesseth et~al.(2018)Naesseth, Linderman, Ranganath, and
  Blei]{Naesseth2018}
Naesseth, C., Linderman, S.~W., Ranganath, R., and Blei, D.~M.
\newblock Variational sequential {M}onte {C}arlo.
\newblock In \emph{Artificial Intelligence and Statistics}, 2018.

\bibitem[Neal(2011)]{Neal2011}
Neal, R.~M.
\newblock {MCMC} using {H}amiltonian dynamics.
\newblock In Brooks, S., Gelman, A., Jones, G.~L., and Meng, X.-L. (eds.),
  \emph{Handbook of Markov Chain Monte Carlo}. Chapman and Hall/CRC, 2011.

\bibitem[Nowozin et~al.(2016)Nowozin, Cseke, and Tomioka]{Nowozin2016}
Nowozin, S., Cseke, B., and Tomioka, R.
\newblock {f-GAN:} training generative neural samplers using variational
  divergence minimization.
\newblock In \emph{Advances in Neural Information Processing Systems}, 2016.

\bibitem[Paisley et~al.(2012)Paisley, Blei, and Jordan]{Paisley2012}
Paisley, J.~W., Blei, D.~M., and Jordan, M.~I.
\newblock Variational {B}ayesian inference with stochastic search.
\newblock In \emph{International Conference on Machine Learning}, 2012.

\bibitem[Papamakarios et~al.(2017)Papamakarios, Murray, and
  Pavlakou]{Papamakarios2017}
Papamakarios, G., Murray, I., and Pavlakou, T.
\newblock Masked autoregressive flow for density estimation.
\newblock In \emph{Advances in Neural Information Processing Systems}, 2017.

\bibitem[Ranganath et~al.(2014)Ranganath, Gerrish, and Blei]{Ranganath2014}
Ranganath, R., Gerrish, S., and Blei, D.~M.
\newblock Black box variational inference.
\newblock In \emph{Artificial Intelligence and Statistics}, 2014.

\bibitem[Ranganath et~al.(2016)Ranganath, Tran, and Blei]{Ranganath2016}
Ranganath, R., Tran, D., and Blei, D.~M.
\newblock Hierarchical variational models.
\newblock In \emph{International Conference on Machine Learning}, 2016.

\bibitem[Reyes-Ortiz et~al.(2016)Reyes-Ortiz, Oneto, Sam\`{a}, Parra, and
  Anguita]{ReyesOrtiz2016}
Reyes-Ortiz, J.~L., Oneto, L., Sam\`{a}, A., Parra, X., and Anguita, D.
\newblock Transition-aware human activity recognition using smartphones.
\newblock \emph{Neurocomputing}, 171\penalty0 (C):\penalty0 754--767, jan 2016.

\bibitem[Rezende \& Mohamed(2015)Rezende and Mohamed]{Rezende2015}
Rezende, D.~J. and Mohamed, S.
\newblock Variational inference with normalizing flows.
\newblock In \emph{International Conference on Machine Learning}, 2015.

\bibitem[Rezende et~al.(2014)Rezende, Mohamed, and Wierstra]{Rezende2014}
Rezende, D.~J., Mohamed, S., and Wierstra, D.
\newblock Stochastic backpropagation and approximate inference in deep
  generative models.
\newblock In \emph{International Conference on Machine Learning}, 2014.

\bibitem[Robert \& Casella(2005)Robert and Casella]{Robert2005}
Robert, C.~P. and Casella, G.
\newblock \emph{{M}onte {C}arlo Statistical Methods ({S}pringer Texts in
  {S}tatistics)}.
\newblock Springer-Verlag New York, Inc., Secaucus, NJ, USA, 2005.

\bibitem[Roeder et~al.(2017)Roeder, Wu, and Duvenaud]{Roeder2017}
Roeder, G., Wu, Y., and Duvenaud, D.
\newblock Sticking the landing: Simple, lower-variance gradient estimators for
  variational inference.
\newblock In \emph{Advances in Neural Information Processing Systems}, 2017.

\bibitem[Salakhutdinov \& Murray(2008)Salakhutdinov and
  Murray]{Salakhutdinov2008_quantitative}
Salakhutdinov, R. and Murray, I.
\newblock On the quantitative analysis of deep belief networks.
\newblock In \emph{International Conference on Machine Learning}, 2008.

\bibitem[Salimans \& Knowles(2013)Salimans and Knowles]{Salimans2013}
Salimans, T. and Knowles, D.~A.
\newblock Fixed-form variational posterior approximation through stochastic
  linear regression.
\newblock \emph{Bayesian Analysis}, 8\penalty0 (4):\penalty0 837--882, 2013.

\bibitem[Salimans et~al.(2015)Salimans, Kingma, and Welling]{Salimans2015}
Salimans, T., Kingma, D.~P., and Welling, M.
\newblock Markov chain {M}onte {C}arlo and variational inference: Bridging the
  gap.
\newblock In \emph{International Conference on Machine Learning}, 2015.

\bibitem[Saul \& Jordan(1996)Saul and Jordan]{Saul1996}
Saul, L.~K. and Jordan, M.~I.
\newblock Exploiting tractable substructures in intractable networks.
\newblock In \emph{Advances in Neural Information Processing Systems}, 1996.

\bibitem[Shi et~al.(2018{\natexlab{a}})Shi, Sun, and Zhu]{Shi2018}
Shi, J., Sun, S., and Zhu, J.
\newblock Kernel implicit variational inference.
\newblock In \emph{International Conference on Learning Representations},
  2018{\natexlab{a}}.

\bibitem[Shi et~al.(2018{\natexlab{b}})Shi, Sun, and Zhu]{Shi2018spectral}
Shi, J., Sun, S., and Zhu, J.
\newblock A spectral approach to gradient estimation for implicit
  distributions.
\newblock In \emph{International Conference on Machine Learning},
  2018{\natexlab{b}}.

\bibitem[Sugiyama et~al.(2012)Sugiyama, Suzuki, and Kanamori]{Sugiyama2012}
Sugiyama, M., Suzuki, T., and Kanamori, T.
\newblock \emph{Density ratio estimation in machine learning}.
\newblock Cambridge University Press, 2012.

\bibitem[Tieleman \& Hinton(2012)Tieleman and Hinton]{Tieleman2012}
Tieleman, T. and Hinton, G.
\newblock Lecture 6.5-{RMSPROP}: Divide the gradient by a running average of
  its recent magnitude.
\newblock Coursera: Neural Networks for Machine Learning, 4, 2012.

\bibitem[Titsias \& L\'{a}zaro-Gredilla(2011)Titsias and
  L\'{a}zaro-Gredilla]{Titsias2011}
Titsias, M.~K. and L\'{a}zaro-Gredilla, M.
\newblock Spike and slab variational inference for multi-task and multiple
  kernel learning.
\newblock In \emph{Advances in Neural Information Processing Systems}, 2011.

\bibitem[Titsias \& L\'{a}zaro-Gredilla(2014)Titsias and
  L\'{a}zaro-Gredilla]{Titsias2014_doubly}
Titsias, M.~K. and L\'{a}zaro-Gredilla, M.
\newblock Doubly stochastic variational {B}ayes for non-conjugate inference.
\newblock In \emph{International Conference on Machine Learning}, 2014.

\bibitem[Tomczak \& Welling(2016)Tomczak and Welling]{Tomczak2016}
Tomczak, J.~M. and Welling, M.
\newblock Improving variational auto-encoders using convex combination linear
  inverse autoregressive flow.
\newblock In \emph{arXiv:1706.02326}, 2016.

\bibitem[Tomczak \& Welling(2017)Tomczak and Welling]{Tomczak2017}
Tomczak, J.~M. and Welling, M.
\newblock Improving variational auto-encoders using {H}ouseholder flow.
\newblock In \emph{arXiv:1611.09630}, 2017.

\bibitem[Tran et~al.(2015)Tran, Blei, and Airoldi]{Tran2015}
Tran, D., Blei, D.~M., and Airoldi, E.~M.
\newblock Copula variational inference.
\newblock In \emph{Advances in Neural Information Processing Systems}, 2015.

\bibitem[Tran et~al.(2016)Tran, Ranganath, and Blei]{Tran2016}
Tran, D., Ranganath, R., and Blei, D.~M.
\newblock Variational {G}aussian processes.
\newblock In \emph{International Conference on Learning Representations}, 2016.

\bibitem[Tran et~al.(2017)Tran, Ranganath, and Blei]{Tran2017}
Tran, D., Ranganath, R., and Blei, D.~M.
\newblock Hierarchical implicit models and likelihood-free variational
  inference.
\newblock In \emph{Advances in Neural Information Processing Systems}, 2017.

\bibitem[Wainwright \& Jordan(2008)Wainwright and Jordan]{Wainwright2008}
Wainwright, M.~J. and Jordan, M.~I.
\newblock Graphical models, exponential families, and variational inference.
\newblock \emph{Foundations and Trends in Machine Learning}, 1\penalty0
  (1--2):\penalty0 1--305, January 2008.

\bibitem[Xiao et~al.(2017)Xiao, Rasul, and Vollgraf]{Xiao2017}
Xiao, H., Rasul, K., and Vollgraf, R.
\newblock Fashion-{MNIST}: A novel image dataset for benchmarking machine
  learning algorithms.
\newblock In \emph{arXiv:1708.07747}, 2017.

\bibitem[Yin \& Zhou(2018)Yin and Zhou]{Yin2018}
Yin, M. and Zhou, M.
\newblock Semi-implicit variational inference.
\newblock In \emph{International Conference on Machine Learning}, volume~80 of
  \emph{Proceedings of Machine Learning Research}, pp.\  5660--5669. PMLR,
  2018.

\bibitem[Zhang et~al.(2017)Zhang, B\"{u}tepage, Kjellstr\"{o}m, and
  Mandt]{Zhang2017}
Zhang, C., B\"{u}tepage, J., Kjellstr\"{o}m, H., and Mandt, S.
\newblock Advances in variational inference.
\newblock \emph{arXiv:1711.05597}, 2017.

\end{thebibliography}
\bibliographystyle{icml2018} 

\end{document}